\documentclass[conference]{IEEEtran}
\IEEEoverridecommandlockouts
\usepackage{mathtools}
\usepackage{amssymb}
\newcommand\NN{\textup{NN}}
\newcommand\softmax{\mathrm{Softmax}}
\newcommand\softplus{\mathrm{Softplus}}
\newcommand\mean{\mathrm{mean}}
\newcommand\nmax{\mathrm{max}}

\DeclareMathOperator*{\argmax}{arg\,max}
\usepackage{lmodern}
\usepackage{siunitx}
\usepackage{booktabs}
\usepackage{etoolbox}
\usepackage[utf8]{inputenc}
\usepackage[english]{babel}
\usepackage{subcaption}
\usepackage[ruled,vlined]{algorithm2e}

\SetCommentSty{mycommfont}
\usepackage{xcolor}
\usepackage{multirow}
\usepackage{makecell}
\usepackage{bbding}
\usepackage{pifont}
\usepackage{wasysym}
\usepackage{amssymb}
\usepackage{xcolor}

\newcommand{\cmark}{\ding{51}}%
\newcommand{\xmark}{\ding{55}}%

\usepackage[breaklinks=true]{hyperref}

\def\BibTeX{{\rm B\kern-.05em{\sc i\kern-.025em b}\kern-.08em
    T\kern-.1667em\lower.7ex\hbox{E}\kern-.125emX}}

\newrobustcmd{\B}{\bfseries}

\begin{document}
\title{Learning Point Processes using Recurrent Graph Network}
\author{Saurabh~Dash \\ \href{mailto:saurabhdash@gatech.edu}{saurabhdash@gatech.edu} 
   \and Xueyuan~She \\ \href{mailto:xshe6@gatech.edu}{xshe6@gatech.edu}
   \and Saibal~Mukhopadhyay \\
   \href{mailto:saibal@ece.gatech.edu}{saibal@ece.gatech.edu}}
\maketitle

\begin{abstract}
We present a novel Recurrent Graph Network (RGN) approach for predicting discrete marked event sequences by learning the underlying complex stochastic process. Using the framework of Point Processes, we interpret a marked discrete event sequence as the superposition of different sequences each of a unique type. The nodes of the Graph Network use LSTM to incorporate past information whereas a Graph Attention Network (GAT Network) introduces strong inductive biases to capture the interaction between these different types of events. By changing the self-attention mechanism from attending over past events to attending over event types, we obtain a reduction in time and space complexity from $\mathcal{O}(N^2)$ (total number of events) to $\mathcal{O}(|\mathcal{Y}|^2)$ (number of event types). Experiments show that the proposed approach improves performance in log-likelihood, prediction and goodness-of-fit tasks with lower time and space complexity compared to state-of-the art Transformer based architectures. 
\end{abstract}

\section{Introduction}
Discrete event sequences are ubiquitous around us - hospital visits, tweets, financial transactions, earthquakes, human activity. These sequences are primarily characterized by the event timestamps however, they can also contain markers and other additional information like type and location. Often these sparse observations have complex hidden dynamics responsible for them and when observed thorough the lens of Point Processes, these sequences can be interpreted as realization of an underlying stochastic process \cite{Cox2018}. This underlying point process can be a simple stochastic process like a Poisson process or could be doubly stochastic and have dependencies on both time and history like the Hawkes process \cite{hawkes}.

Traditionally these processes have been described using simplified historical dependencies. The Hawkes process assumes additive mutual excitation however, these underlying assumptions are not valid for practical real world data. Recent works seek to leverage the advances in deep learning to relax these restrictions by defining a rich, flexible family of models to incorporate historical information \cite{rmtpp, nhp, thp, sahp, Shchur2020Intensity-Free}. Recurrent Neural Network (RNN) \cite{rmtpp, nhp, Shchur2020Intensity-Free} and Transformer based \cite{thp, sahp} architectures have been explored to capture historical dependencies to better predict the time and type of future events compared to a pre-speciﬁed Hawkes Process. However, shallow RNNs only capture simple state transitions and multi-layer RNNs are diﬃcult to train. On the other hand, Transformer based approaches are highly expressive and interpretatable but the multi-headed self-attention mechanism consumes exorbitant amounts of memory for large sequences.

Our novel approach simultaneously solves the problems faced by the above approaches. An entire discrete event sequence with marks can be interpreted as a superposition of multiple sequences each only containing events of a single unique marker. We process the evolution of each of these unique event type sequences with an LSTM \cite{lstm}. These different LSTM hidden states are then used to generate a dynamic relational graph and model complex dependencies between event types using a Graph Attention Network \cite{gat} as shown in figure \ref{updates}. {A key innovation of our architecture lies in attending over event types instead of all past events}. This leads to reduction in space (memory) and time (number of computations) complexity as instead of calculating attention over all past inputs - which can get prohibitively expensive as the sequences get longer $\mathcal{O}(N^2)$, we attend over event types to obtain a time-varying attention matrix with constant time and space complexity in the order of $\mathcal{O}(|\mathcal{Y}|^2)$, independent of the sequence length. Additionally, this time-varying event attention matrix helps with interpretability as it encodes the evolution of the underlying relational graph between event types as new events are observed in the data stream. To the best of our knowledge, this is the first time a Recurrent Graph Network using expressive attention mechanism to encode a dynamic relational graph is used for learning Point Processes. \textcolor{black}{The advantages of the proposed method over previous approaches is summarized in table \ref{relatedwork}.}

The major contributions of this paper are summarised as follows: 
\begin{itemize}
    \item We present a Recurrent Graph Network (RGN) approach to learn point processes that uses strong inductive biases to better encode historical information to model the conditional intensity function.
    \item We show that the proposed model performs better in log-likelihood and predictive tasks for multiple datasets.
    \item We show that this formulation allows for interpretability of the underlying temporally evolving Relational Graph between event types.
    \item We show that RGN has lower computational complexity and activation footprint than state-of-the art Transformer based approaches leading to computation and memory savings.
\end{itemize}
We perform experiments on standard point process datasets like - Retweets \cite{Zhao_2015}, Financial Transactions \cite{rmtpp}, StackOverflow \cite{snapnets} and MIMIC-II \cite{Johnson2016}. In addition to these, we also introduce a new dataset for benchmarking point process models extracted from StarCraft II game replays. The underlying complex stochastic process that generates this data is the high level strategy that the players adopt over the course of the game. The Starcraft II dataset can be found here\footnote{https://figshare.com/s/c028296e953788b25599} and the supplementary materials can be found here\footnote{https://figshare.com/s/7a10a6aa2c0752f2cca3}.

Our experiments show that RGN improves over state-of-the art Transformer Hawkes Process (THP) on log-likelihood by upto $8\%$, event type prediction accuracy by upto $4\%$ and event time error by upto $13\%$.

\begin{table}[t]
\caption{Comparison with previous approaches}
\label{relatedwork}
\centering
\begin{tabular}{c | c c c }
\hline
Model & \makecell{Stack \\ Multiple \\ Layers} & \makecell{Small \\ Activation \\ Footprint} &  \makecell{Model \\ Interpretability}\\
\hline
RMTPP \cite{rmtpp} & \color{red} \xmark & \color{green} \cmark & \color{red} \xmark  \\
NHP \cite{nhp} & \color{red} \xmark & \color{green} \cmark & \color{red} \xmark  \\
THP \cite{thp} & \color{green} \cmark & \color{red} \xmark & \color{green} \cmark \\
\B{This work} & \color{green} \cmark & \color{green} \cmark & \color{green} \cmark \\
\hline
\end{tabular}
\end{table}

\section{Related Work}
Du et al. \cite{rmtpp} propose using an RNN (RMTPP) to encode the history to a hidden state $\mathbf{h}_t$ that is used to define the conditional intensity $\lambda^*(t) = \exp(\boldsymbol{v}^T h_i + w(t - t_i) + b)$. Mei \& Eisner \cite{nhp} proposed a novel RNN architecture - Continuous-time LSTM (CLSTM) which modifies the LSTM cell to incorporate a decay $\delta$ when an event is not observed to a steady state $\boldsymbol{\bar{c}}$. Since CLSTM work in continuous time, there is no need to explicitly feed the timestamps of the events like RMTPP. However both models only contain shallow RNNs which only capture simple state transitions \cite{279181} as multi-layer RNNs are difficult to train \cite{pmlr-v28-pascanu13} due to vanishing / exploding gradients. To alleviate these concerns Zuo et al. \cite{thp} and Zhang et al. \cite{sahp} make use of multi-headed self attention \cite{mhsa} based Transformer architectures to better distill out historical influence. This allows using multi-layer architectures to model complex interactions compared to RNN based architectures, however the $\mathcal{O}(N^2)$ cost of self attention computation leads to prohibitively large memory footprint for longer sequences. Even if a single sequence in a mini-batch is long, the other sequences have to be zero padded to match the longest length for batch processing, leading to a lot of wasted memory and redundant computation.
In contrast, Schur et al. \cite{Shchur2020Intensity-Free} draw upon advances in Normalizing Flows \cite{nf} and propose a mixture distribution to define a model using the conditional density function instead of the conditional intensity function. This allows for the conditional density function to be multi-modal allowing for much more complex distributions; however, the historical influence is again encoded with a simple RNN which leads to degraded performance due to lack to rich historical information.

\textcolor{black}{Chang et al. \cite{10.1145/3340531.3411946} propose a dynamic message passing neural network named TDIG-MPNN which aims to learn continuous-time dynamic embeddings from sequences of interaction data. Although our proposed network also benefits from learning dynamic graph embeddings, the end goal of this work is to learn the underlying conditional intensity function that is responsible for observed data and not the time-varying interaction graph between different entities. Moreover, the datasets under consideration here do not contain any interaction data. The work closest to the proposed approach is ARNPP-GAT \cite{closest}. ARNPP-GAT divides the representations of users (marks) into two categories: long-term representation which is modelled using Graph Attention Layer (GAT Layer) and short-term representation which is modelled using a Gated Recurrent Unit (GRU). Although we use similar blocks, the architecture of the model is drastically different. ARNPP-GAT leverages GAT layer to model a time-independent interaction between users (marks) while the historical context is modelled by a GRU. These embeddings are then combined for predicting the time and location of next event. On the other hand, RGN uses GAT layers at every time-step to model a dynamic relational graph between different event types. Moreover, we also use an additional log-likelihood loss term which encourages the model to learn the underlying stochastic process better.}

\section{Background}
\subsection{Point Processes}
A temporal point process is a stochastic process composed of a time series of events that occur instantaneously in continuous time \cite{Cox2018, daley}. The realization of this point process is a set of strictly increasing arrival times $\tau = \{t_1,..., t_N\}$. We are interested in a Marked Point Process where each event also has an associated marker $y \in \mathcal{Y}$. In real world data, event probabilities can be affected by past events, thus the probability of an event occurring in an infinitesimally small window $[t, t + dt)$ is called the conditional intensity $\lambda^*(t) \coloneqq \lambda \left(t| \mathcal{H}_t\right)$. $\mathcal{H}_t$ denotes the history $\{(t', y') \in \tau| t' < t\}$. The $*$ symbol represents the dependence on history. Given the conditional intensity function, the conditional density function can be obtained as shown in Eq. \ref{pdf} below \cite{rasmussen2018lecture}. 

\begin{equation}
    \label{pdf}
    f^*(t) = \lambda^*(t) \: \exp\left({- \int_{t_n}^{t} \lambda^*(s) ds}\right)
\end{equation}

    

\subsection{Recurrent Networks}
Recurrent Networks are autoregressive models that can incorporate past information along with the present input for prediction. RNNs and its modern variants LSTM \cite{lstm} and GRU \cite{chung2014empirical} have shown remarkable success in sequence modelling tasks like - language modelling \cite{seq2seq, DBLP:journals/corr/ChoMGBSB14}, video processing \cite{videolstm} and dynamical systems \cite{saha2020physicsincorporated}. At each time step, an input $\boldsymbol{i}_t$ is fed into the model with the past internal state information $\boldsymbol{h}_t$. The internal state is updated and used to predict the output $\boldsymbol{o}_t$. 


\subsection{Graph Networks}
Graph Networks (GN) \cite{battaglia2018relational} describe a class of functions for relational reasoning over graph structured data. These take in a graph structured data denoted by the 3-tuple $(\mathbf{u}, V, E)$. $\mathbf{u}$ denotes the global attributes, $V \coloneqq \{\mathbf{v}_i\}_{i=1}^{N^v}$ denotes the set of node attributes and $E \coloneqq \{ (\mathbf{e}_i, r_k, s_k)\}_{i=1}^{N^e}$ denotes the set of edges of a graph where $\boldsymbol{e}_k$ is the edge attribute, $r_k$ is the receiver node and $s_k$ is the sender node.

A full GN block enables flexible representations in terms of representation of attributes and in terms of structure of the graph and can be used to describe a wide variety of architectures \cite{gat, mhsa}. The main unit of computation is the GN block which performs graph-to-graph operations. The GN block takes a graph structure as an input, performs operations over the graph and outputs a graph.

A full GN block contains three update functions $\phi$ and three aggregate functions $\rho$ \cite{battaglia2018relational}:
\begin{align}
    \boldsymbol{e}'_k & = \phi^e(\boldsymbol{e}_k, \boldsymbol{v}_{r_k}, \boldsymbol{v}_{s_k}, \boldsymbol{u}) \\
    \boldsymbol{v}'_i&  = \phi^v(\bar{\boldsymbol{e}}'_i, \boldsymbol{v}_i, \boldsymbol{u})  \\
    \boldsymbol{u}' & = \phi^u(\bar{\boldsymbol{e}}', \bar{\boldsymbol{v}}', \boldsymbol{u}) \\
    \bar{\boldsymbol{e}}'_i & = \rho^{e \rightarrow v}(E'_i) \\
    \bar{\boldsymbol{e}}' & = \rho^{e \rightarrow u}(E') \\
    \bar{\boldsymbol{v}}' & = \rho^{v \rightarrow u}(V')
\end{align}
where, $E'_i \coloneqq \{ (\boldsymbol{e}'_k, r_k, s_k)\}_{r_k = i, k = 1}^{N_e}$ is the set of all updated edge attributes that have node $i$ as the receiver, $V' \coloneqq \{\boldsymbol{v}'_i\}_{i=1}^{N_v}$ is the set of updated node attributes and $E' \coloneqq \bigcup_{i} E'_i = \{ (\boldsymbol{e}_k, r_k, s_k)\}_{k=1}^{N_e}$ is the set of updated edge attributes. The update functions $\phi$ are shared across the nodes and edges allowing for a reduction in parameter count and encouraging a form of combinatorial generalization \cite{battaglia2018relational}. These can be any arbitrary functions or more generally parameterized using a Neural Network architecture like - Multi Layer Perceptron (MLP) or incorporate past information using a Recurrent Neural Network (RNN). On the other hand, $\rho$ are permutation invariant aggregation functions which take a set of inputs and reduce them to a single aggregate element, for example $\nmax$ or $\mean$.

\section{Proposed Architecture}
\begin{figure*}[h]
\centering
\includegraphics[width=0.9\textwidth, trim={6cm 13cm 18.7cm 6.8cm}, clip]{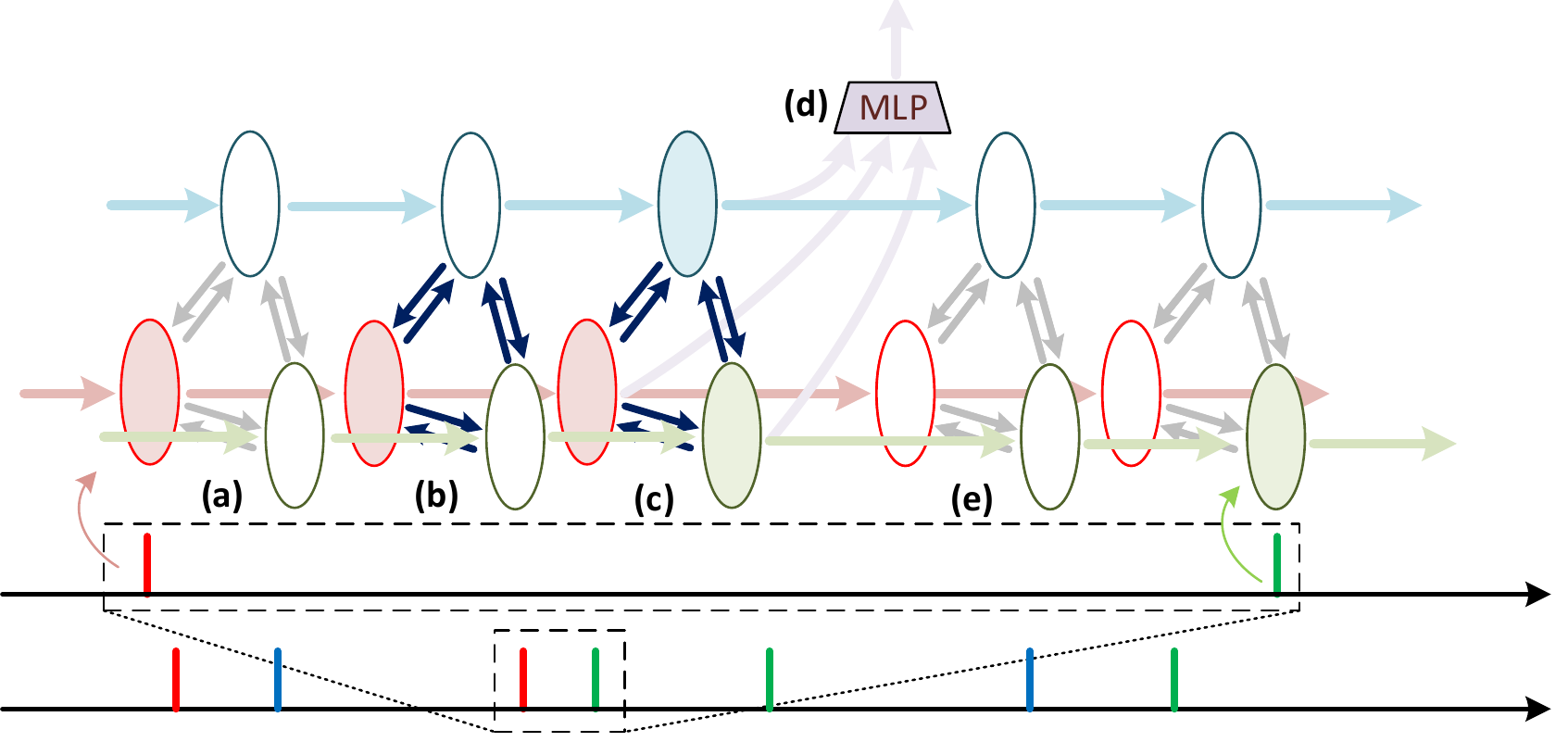}
\caption{Updates in Recurrent Graph Network: (a) \textit{Input-to-Graph}: The model observes an input event, using the past node attributes, updates event node.(b)\textit{Edge Update}: Based on the updated node, edge updates are calculated using attention mechanism.(c)\textit{Node Update}: Nodes are updated using edge aggregation.(d)\textit{Global update}: Updated node attributes are passed through an MLP to predict the conditional intensity, type of next event, and time of next event.(e)\textit{Propagation}: Current node attributes are again fed back to the model as past history when a new event is observed. The red and green lines are the observed events sequentially appearing in time.}
\label{updates}
\end{figure*}

\subsection{Recurrent Graph Network for Learning Point Processes}
An entire discrete event sequence with marks can be interpreted as a superposition of multiple sequences each only containing events of a single unique marker. Using the input sequence, we build a fully connected graph with $|\mathcal{Y}|$ nodes, each containing a latent embedding for a specific type marker which allows us to understand how different event types influence future events. As we are building a relational graph from non-graph structured inputs, the relational information in the graph is not explicit and needs to be inferred. This relational graph has node attributes and edge weights evolving with time as new events are observed. Thus in a sequence $\mathcal{S}_i$, with $N_T$ events, we can interpret the relational graph undergoing $N_T$ transitions which can be processed naturally using a Recurrent Graph Network. We are interested in predicting the conditional intensity $\lambda^*(t)$ which is a global property, which makes the Recurrent Graph Network a graph focused Graph Network \cite{battaglia2018relational}. 

At each event $(t_i, y_i)$ in an event sequence $\mathcal{S}$, the inputs to the model are the previous node attributes $\{ \boldsymbol{v}_{t_{i-1}}^{y}\}_{y=1}^{|\mathcal{Y}|}$, $\boldsymbol{v}_{t_{i-1}}^{y} \in \mathbb{R}^{d_v}$ and the input embedding $\boldsymbol{x}_{t_i}$. Once a new event is observed, there are three types of updates that take place:
\subsubsection{Input-to-Graph Update}
Each unique-marked sequence is assigned a designated graph node for processing. When an event of a particular type is observed, we want to first update the node attribute corresponding to this marker. This is performed using an LSTM \cite{lstm}. The LSTM corresponding to node $y_{i}$ updates its node attribute $\boldsymbol{v}^{y_i}_{t_i}$ using the previous node attribute $\boldsymbol{v}^{y_i}_{t_{i-1}}$ and $\boldsymbol{x}_{t_i}$ using the following equations, while the rest of the node attributes remain unchanged in this update.
\begin{align}
        \boldsymbol{f}^{y_i}_{t_i} & = \sigma(W^y_f \boldsymbol{x}_{t_i} + U^y_f \boldsymbol{v}^{y_i}_{t_{i-1}} + \boldsymbol{b}^{y_i}_f) \\
        \boldsymbol{\iota}^{y_i}_{t_i} & = \sigma(W^y_i \boldsymbol{x}_{t_i} + U^y_i \boldsymbol{v}^{y_i}_{t_{i-1}} + \boldsymbol{b}^{y_i}_i) \\
        \boldsymbol{o}^{y_i}_{t_i} & = \sigma(W^y_o \boldsymbol{x}_{t_i} + U^y_o \boldsymbol{v}^{y_i}_{t_{i-1}} + \boldsymbol{b}^{y_i}_o) \\
        \boldsymbol{g}_{t_i}^{y_i} & = \tanh(W^y_g \boldsymbol{x}_{t_i} + U^y_g \boldsymbol{v}^{y_i}_{t_{i-1}} + \boldsymbol{b}^y_g) 
        \\
        \boldsymbol{c}_{t_i}^{y_i} & = \boldsymbol{f}^{y_i}_{t_i} \circ \boldsymbol{c}^{y_i}_{t_{i-1}} + \boldsymbol{\iota}^{y_i}_{t_i} \circ \boldsymbol{g}_{t_i}^{y_i} 
        \\
        \boldsymbol{v}^{y_i}_{t_i} & = \boldsymbol{o}^{y_i}_{t_i} \circ \tanh(\boldsymbol{c}^{y_i}_{t_i})
\end{align}

\subsubsection{Graph-to-Graph Update}
Once the node attribute $\boldsymbol{v}_{t_i}^{y_i}$ corresponding of the observed event type $y_i$ node has been updated, information needs to be propagated to other nodes to update the node attributes. For this purpose, we propose using Graph Attention Network (GAT) \cite{gat} as the attention mechanism can be used to assign edge weights not explicitly present in the relational graph using the node attributes. In this section, we express the Graph Attention Layer \cite{gat} in terms of Graph Network \cite{battaglia2018relational} operations. For notational simplicity, we drop the temporal dependence of node attributes $\boldsymbol{v}_{t_i}^{y}$ at current time $t_i$ and represent them with $\boldsymbol{v}_y$.

\textbf{Edge Update:} The edge update function $\phi^e$ only uses the node attributes of the sender node $s_k$ and receiver node $r_k$ and outputs a scalar $a'_k \in \mathbb{R}^+$, vector $\boldsymbol{b}'_k \in \mathbb{R}^{d_e}$ tuple. $\NN$ represents a multi-layer perceptron (MLP).

\begin{align}
    \boldsymbol{e}'_k & = (a'_k, \boldsymbol{b}'_k) = \phi^e(e_k, \boldsymbol{v}_{r_k}, \boldsymbol{v}_{s_k}, \boldsymbol{u}) \coloneqq f^e(\boldsymbol{v}_{r_k}, \boldsymbol{v}_{s_k})   \\
    a'_k &  = \exp(\NN_{\alpha'}([\NN_{\alpha}(\boldsymbol{v}_{r_k}), \NN_{\alpha}(\boldsymbol{v}_{s_k})])) \\
    b'_k & = \NN_\beta (\boldsymbol{v}_{s_k})
\end{align}

\textbf{Edge Aggregation:} $a'_k$ and $\boldsymbol{b}'_k$ are then used by the aggregation function $\rho^{e \rightarrow v}$ to aggregate the edges which have node $i$ as the receiver into $\bar{\boldsymbol{e}}'_i \in \mathbb{R}^{d_e}$. The scalar terms $a'_k$ are normalized to obtain the attention scores, which are used as the weights for weighted element-wise summation of $\boldsymbol{b}'_k$.

\begin{align}
    \bar{\boldsymbol{e}}'_i = \rho^{e \rightarrow v}(E'_i) \coloneqq \frac{1}{\sum_{k: r_k = i} a'_k} \sum_{k: r_k = i} a'_k \boldsymbol{b}'_k
\end{align}

\textbf{Multi-headed Attention: } The edge update and aggregation steps can be performed independently $N_h$ times simultaneously on the same input. We observed this improves the performance of the model similar to the results observed by Veli\v{c}kovi\'{c} et al. \cite{gat} as it allows the network to jointly attend to input projections in various representation subspaces \cite{mhsa}. All the $N_h$ different edge updates $\bar{\boldsymbol{e}}'^h_{i}$ corresponding to different attention heads are concatenated into one single vector $\Tilde{\boldsymbol{e}}'_i \in \mathbb{R}^{N_h \times d_e}$. $
    \Tilde{\boldsymbol{e}}'_i = [\bar{\boldsymbol{e}}'^1_{i},...,\bar{\boldsymbol{e}}'^{N_h}_{i}]$

\textbf{Node Update: } Node attributes $\boldsymbol{v}_i \in \mathbb{R}^{d_v}$ are updated by passing the multi-headed edge aggregation $\Tilde{\boldsymbol{e}}'_i$ through an MLP $\NN_v$. 

\begin{align}
    \boldsymbol{v}'_i = \phi^v(\bar{\boldsymbol{e}}'_i, \boldsymbol{v}_i, \boldsymbol{u}) \coloneqq f^v(\{\bar{\boldsymbol{e}} '^h_i\}_{h=1}^{N_h}) = \NN_v(\Tilde{\boldsymbol{e}}'_i)
\end{align}

\subsubsection{Global Update}
Once the node attributes are updated, we concatenate all the node attributes and pass it through an MLP $\NN_u$ to obtain a global attribute $\boldsymbol{u}' \in \mathbb{R}^{d_u}$ which is used to predict - conditional intensity for all event types $\lambda^*_{y}\in \mathbb{R}^{|\mathcal{Y}|}_{+}$, type of next event $\hat{y}_{i+1} \in \mathcal{Y}$ and time of the next event $\hat{t}_{i+1} \in (t_i, \infty)$. This global attribute represents the history $\mathcal{H}_{t}$ upto time $t_i$. We add $t_i$ in subscript to convey the temporal dependence of this global attribute. 
\begin{align}
    \boldsymbol{u}_{t_i} = \boldsymbol{u}' & = \phi^u(\bar{\boldsymbol{e}}, \bar{\boldsymbol{v}}, \boldsymbol{u}) \coloneqq \NN_u([\boldsymbol{v}'_1,..,\boldsymbol{v}'_{|\mathcal{Y}|}]) \label{hiddvec} \\
    \hat{y}_{i+1} & = \argmax_{y} \softmax(\NN_y(\boldsymbol{u}_{t_i})) \\
    \hat{t}_{i+1} & = \NN_t(\boldsymbol{u}_{t_i})
\end{align}

We define our model in terms of conditional intensity due to its simplicity. As it is not a probability density, the only restriction is that it has to be non-negative. Moreover, it need not sum up to $1$ unlike the conditional density. The conditional intensity of a Hawkes process is defined at every point $t \in \mathbb{R}^+$, however, equation \ref{hiddvec} only outputs the hidden representations at the timestamps in the observed sequence. To interpolate the conditional intensity $\lambda^*$ to the time when an event does not occur, we use the expression proposed by Zuo et al. \cite{thp}.

\begin{align}
    \lambda^*(t) & = \sum_{y=1}^{|\mathcal{Y}|} \lambda_y^{*}(t) \\
    \lambda^{*}_y(t) & = \softplus(\alpha_y \frac{t - t_i}{t_i} +  \NN_\lambda(\boldsymbol{u}_{t_i})_y + \beta_y) \label{softplus}
\end{align}

The softplus function guarantees that the $\lambda^{*}_y$ are non-negative. 
The first term of equation \ref{softplus} represents the contribution of the current event time towards the future and $\alpha_y$ is a hyperparameter. 
The second term is the history term that encodes how the history $\mathcal{H}_t$ of observed events influence the conditional intensity of a certain event type.
The last term $\beta_y$ incorporates the base intensity of the point process in the absence of events.

Updated node attributes $\{\boldsymbol{v}'_y\}_{y =1}^{|\mathcal{Y}|}$ now become the previous node attributes $\{ \boldsymbol{v}_{t_{i}}^{y}\}_{y=1}^{|\mathcal{Y}|}$ when the next event $(t_{i+1}, y_{i+1})$ is processed by the model. The entire information flow is illustrated in Fig. \ref{updates}.

\subsection{Input Embedding} \label{inpemb}
Although RNN allows for a natural ordering in the processing of inputs, as the arrival of the events is not uniform, temporal information needs to be explicitly fed to the model. Directly feeding the time $t_i$ causes issues as The input value increases unbounded or the model may not see any sequence with one specific length which would hurt generalization. To overcome this issue, we follow the positional embedding proposed by Vaswani et al. \cite{mhsa}.
The trigonometric functions ensure that the input embedding $\boldsymbol{x}_{t_i} \in \mathbb{R}^{d_x}$ is bounded and deterministic which enable generalization to longer sequences of unseen lengths.

\subsection{Learning Objectives}
\textbf{Log-Likelihood: } We use maximum likelihood estimation (MLE) to learn the parameters of the model. For a sequence $\mathcal{S}_i \coloneqq \{(t_j, y_j) \;| \; t_j \leq T ,\; j \in \{1,...,L_i\} \}$, the log-likelihood of observing the sequence is given by: 
\begin{align}
    \ell(\mathcal{S}_i) = \sum_{j = 1}^{L_i} \log\lambda^{*}_{y_j}(t_j) - \int_{0}^{T} \lambda^{*}(t) \: dt \label{ll}
\end{align}

The first term in the log-likelihood expression Eq. \ref{ll}, is the log-likelihood of an event occurring at times $t_j$, we would like this term to be as large as possible. Whereas, the second term signifies the log-likelihood of no events occurring in the times $t$ other than $t_j$. We would like this term to be as small as possible.

As the sequences $\mathcal{S}_i$ in the dataset are assumed to be i.i.d, the loss function to be minimized is the negative of the sum of the log-likelihood over all sequences $\mathcal{L}_\lambda = -\sum_{i=1}^{N_\mathcal{D}}\ell(\mathcal{S}_i)$.

The integral $\Lambda = \int_{0}^{T} \lambda^{*}(t) \: dt$ in Eq. \ref{ll} does not have a closed form solution and needs to be numerically approximated. We use the Monte Carlo estimate \cite{mc} given in the supplementary material.


\textbf{Event Prediction: }
We are also interested in predicting the type of the next event, we additionally impose a cross-entropy loss term that penalizes when the model mispredicts the type of the next event. For an event $(t_j, y_j)$, let $\boldsymbol{y}_j \in \mathbb{R}^{|\mathcal{Y}|}$ be the one-hot encoding of the event type $y_j$.Hence the next event prediction loss is given by: $\mathcal{L}_y = - \sum_{i=1}^{N_\mathcal{D}} \sum_{j=2}^{L_i} \boldsymbol{y}_j^T \log(\NN_y(\boldsymbol{u}_{t_j}))$.

\textbf{Time Prediction: }
Apart from event prediction, we also want the predicted time of the next event to be close to the ground truth. Time of the next event is a continuous value that needs to be estimated thus we use an $L2$ penalty to reduce the Mean Squared Error (MSE). The time prediction loss is given by: $
    \mathcal{L}_t = \sum_{i=1}^{N_\mathcal{D}} \sum_{j=2}^{L_i} |t_j - \hat{t}_j|^2$.


\section{Experiments}

\subsection{Datasets}
\textbf{Retweets(RT) \cite{Zhao_2015}: } This dataset contains sequences of tweets where each sample has a sequence of tweets of different users. 

\textbf{StackOverflow(SO) \cite{snapnets}: } The StackOverflow dataset contains sequences of awards that users were awarded for answering questions on the StackOverflow website. The markers for the events are the various different awards.

\textbf{MIMIC-II \cite{Johnson2016}: } MIMIC-II dataset contains the visitation of various patients to a Hospital's ICU. Each sample represents a patient and the events markers are the diagnosis.

\textbf{Financial Transactions \cite{rmtpp}: } This dataset contains the transaction records for multiple stocks on a single day. The different events are buy and sell orders at various timestamps.

\textbf{StarCraft II(SC-II): } We introduce a new dataset for benchmarking various discrete event models. Each sequence is a ``build-order'' which represents a temporally ordered list of various buildings built by the players over the course of a game. The strategy of the players is the underlying complex stochastic process that generates this data. Each sequence in the dataset is an entire Protoss vs Protoss game. Additional details are presented in the supplementary material.


\subsection{Setup}
Here we describe the architectural choices we make for various update function. In $a'_k$, $\NN_\alpha = \NN_\beta$ is simply a linear projection $W \in \mathbb{R}^{d_{v} \times d_e}$, $\NN_{\alpha'}$ is a single fully connected (FC) layer with leakyReLU \cite{gat} non-linearity. $NN_v$ is a linear projection from $\mathbb{R}^{N_h \times d_e}$ to $\mathbb{R}^{d_v}$. $\NN_u$ uses a single FC layer with ReLU non-linearity whereas $NN_y$, $NN_t$ and $NN_\lambda$ are simple linear projections of size $R^{d_u \times |\mathcal{Y}|}$, $\mathbb{R}^{d_u \times 1}$ and $\mathbb{R}^{d_u \times |\mathcal{Y}|}$ respectively.



We are interested in the thee evaluation metrics - Log-likelihood, event class prediction accuracy and event time prediction error. We also evaluate these metrics for - Transformer Hawkes Process (THP) \cite{thp}, Recurrent Marked Temporal Point Process (RMTPP) \cite{rmtpp} and Neural Hawkes Process (NHP)\cite{nhp} for comparison. For a fair comparison, we optimize these models using the same objective as ours and for each evaluation metric we pick the model parameters which lead to the best performance on the validation set. Additional training details are included in the supplementary material.

\begin{figure*}
\centering
\includegraphics[width=0.9\textwidth, trim={3cm 2cm 2.7cm 2cm}, clip]{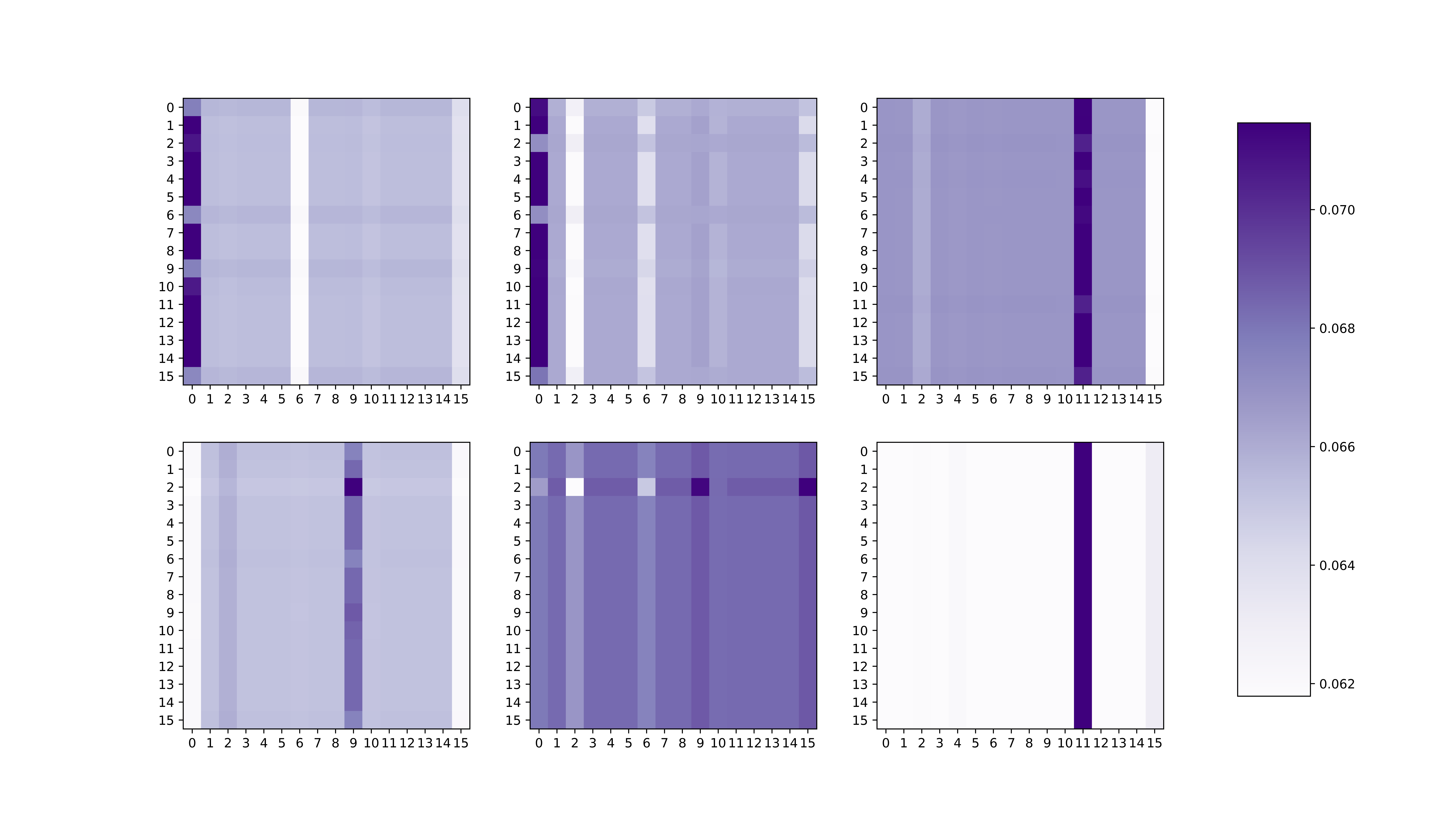}
\caption{Visualization of various attention heads (row) for different timestamps (column)}
\label{attn}
\end{figure*}

\subsection{Likelihood}
\begin{table}[!t]

\addtolength{\tabcolsep}{-4.1pt}

\caption{Log-Likelihood Comparison.}
\label{likelihood}
\centering
\begin{tabular}{c | c c c c c}
\hline
Model & RT & SO &  MIMIC-II & Financial & SC-II \\
\hline
RMTPP & -8.8 & -0.73 & -0.39 & -1.71 & 1.09 \\
NHP & -8.24 & -2.41 & -1.38 & -2.60 & -0.81 \\
THP & -7.80  & -0.73 & 0.86 & -1.53 & 1.04 \\
\B{This work} & \underline{\B{-6.76}} & \underline{\B{-0.56}} & \underline{\B{1.00}} & \underline{\B{-1.26}} & \underline{\B{1.50}} \\
\hline
\end{tabular}
\end{table}

In this section, we use log-likelihood as an evaluation metric as done in previous works \cite{Shchur2020Intensity-Free, sahp, thp, nhp, rmtpp}. A higher log-likelihood score would imply that the model can approximate the conditional intensity function $\lambda^*(t)$ well.
Table \ref{likelihood} shows the log-likelihood results. We can see that RGN beats THP in log-likelihood across all the datasets.

\subsection{Event Type Prediction}
\begin{table}
\addtolength{\tabcolsep}{-4.1pt}
\caption{Event Prediction Accuracy Comparison.}
\label{eventprediction}
\centering
\begin{tabular}{c | c c c c c}
\hline
Model & RT & SO & Financial  &  MIMIC-II & SC-II \\
\hline
RMTPP & 49.89 & 43.49 & 60.29 & 59.91 & 44.25 \\
NHP & 48.82 & 42.81 & 61.20 & 76.21 & 41.14 \\
THP & 50.43 & 43.23 & 61.67 & \underline{80.14} & 43.33 \\
\B{This work} & \underline{\B{54.42}} & \underline{\B{45.46}} & \underline{\B{61.69}} & {\B{77.34}} & \underline{\B{47.31}} \\
\hline
\end{tabular}
\end{table}

We are also interested in the predictive performance of the model to predict the type of next event. To improve the performance of the model on this evaluation metric, we added an extra event type prediction loss $\mathcal{L}_y$ in addition to the standard log-likelihood objective. The results are presented in table \ref{eventprediction}. Unlike the log-likelihood results, we see that although RGN has better predictive performance than THP on most of the datasets, it performs worse on the MIMIC-II dataset. 

\subsection{Event Time Prediction}
To improve the performance of the model on this evaluation metric, we added an extra event time prediction loss $\mathcal{L}_t$ in addition to the standard log-likelihood objective. Table \ref{timeprediction} shows the Root Mean-Squared Error (RMSE) of the proposed approach compared to the baselines. The lower the RMSE, the better the performance of the model. We observe that RGN outperforms THP in four of the five datasets however THP again performs better on the MIMIC-II dataset.

\begin{table}

\addtolength{\tabcolsep}{-4.1pt}

\caption{Time Prediction $L2$ Error Comparison.}
\label{timeprediction}
\centering
\begin{tabular}{c | c c c c c}
\hline
Model & RT & SO &  MIMIC-II & Financial & SC-II \\
\hline
RMTPP & 16899.72  & 144.34 & 3.42 & 26.95 & 1.48 \\
NHP & 17672.54  & 144.72 & 3.17 & 27.88 & 1.51 \\
THP & 16616.13 & 140.20  & \underline{0.87} & 25.75 & 1.39 \\
\B{This work} & \underline{\B{15999.68}} &  \underline{\B{121.50}} & \B{1.026} &  \underline{\B{25.37}} & \underline{\B{1.25}} \\
\hline
\end{tabular}
\end{table}

\subsection{Goodness of Fit}
\begin{figure*}[!h]
\centering
\begin{subfigure}{.25\textwidth}
  \centering
  \includegraphics[width=\linewidth, trim={0.5cm 0.1cm 1.6cm 1.4cm}, clip]{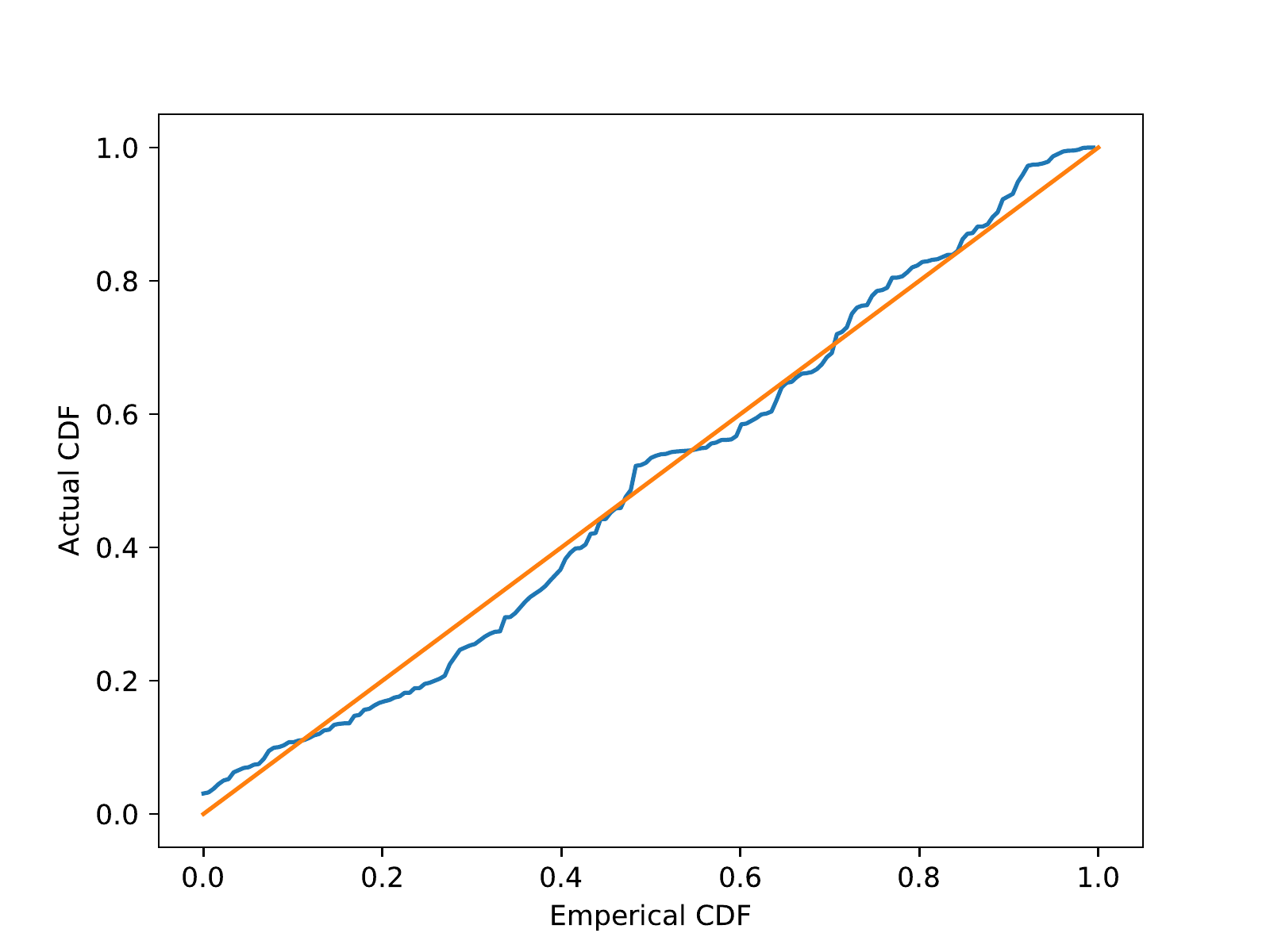}
  \label{fig:sfig1}
\end{subfigure}%
\begin{subfigure}{.25\textwidth}
  \centering
  \includegraphics[width=\linewidth, trim={0.5cm 0.1cm 1.6cm 1.4cm}, clip]{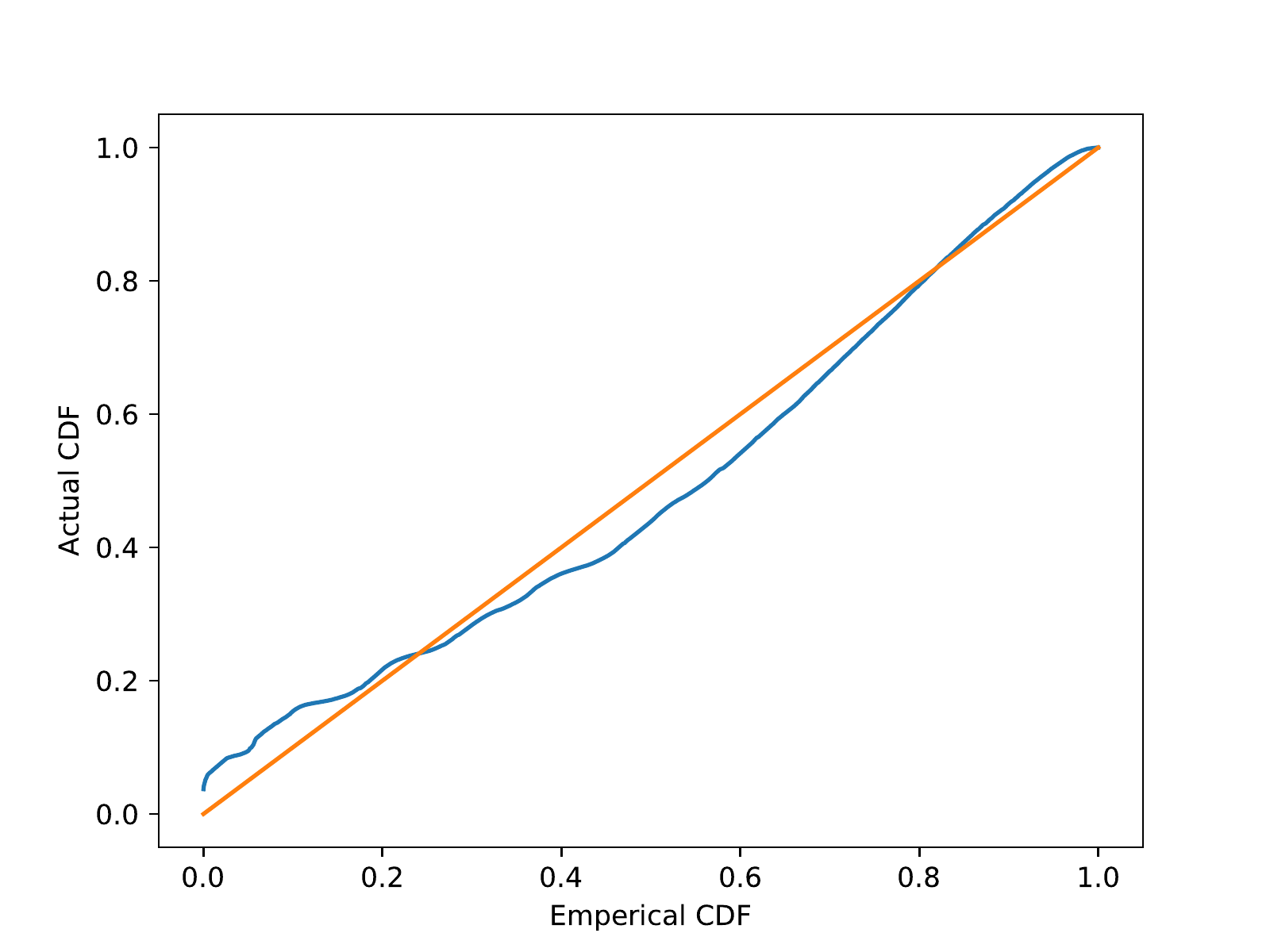}
  \label{fig:sfig2}
\end{subfigure}
\begin{subfigure}{.25\textwidth}
  \centering
  \includegraphics[width=\linewidth, trim={0.5cm 0.1cm 1.6cm 1.4cm}, clip]{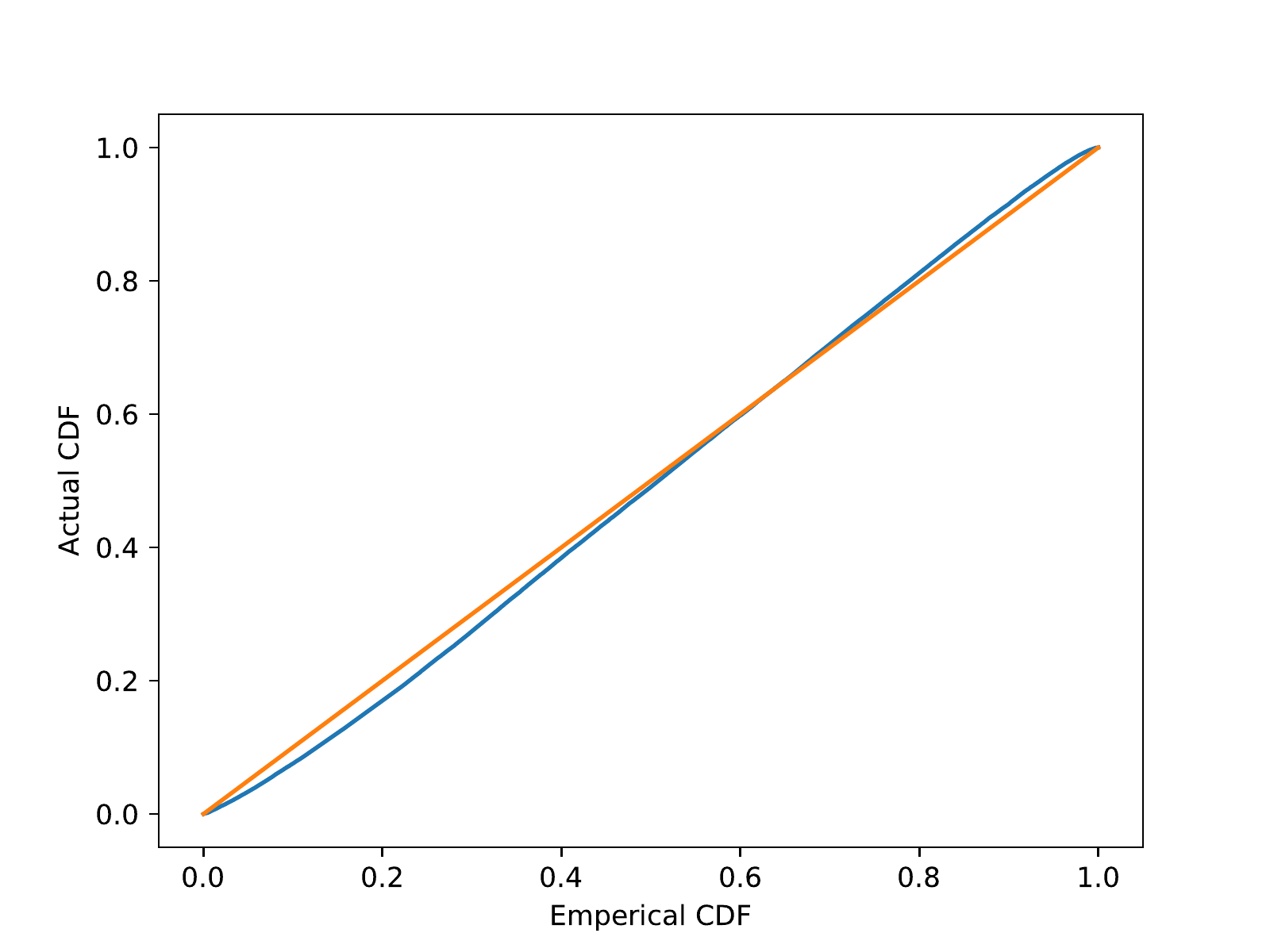}
  \label{fig:sfig1}
\end{subfigure}
\\
\begin{subfigure}{.25\textwidth}
  \centering
  \includegraphics[width=\linewidth, trim={0.5cm 0.1cm 1.6cm 1.4cm}, clip]{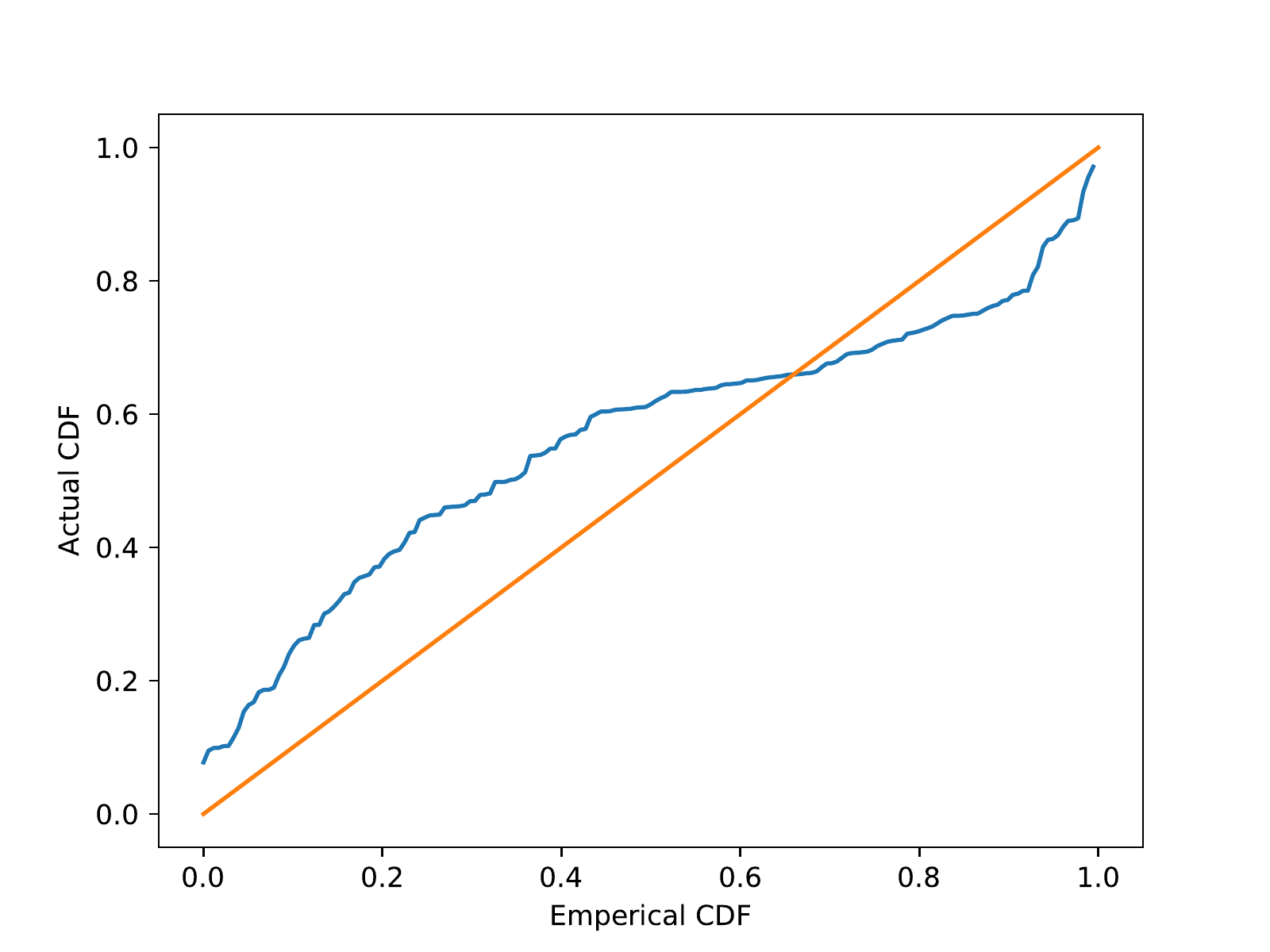}
  \caption{MIMIC-II}
  \label{fig:sfig1}
\end{subfigure}%
\begin{subfigure}{.25\textwidth}
  \centering
  \includegraphics[width=\linewidth, trim={0.5cm 0.1cm 1.6cm 1.4cm}, clip]{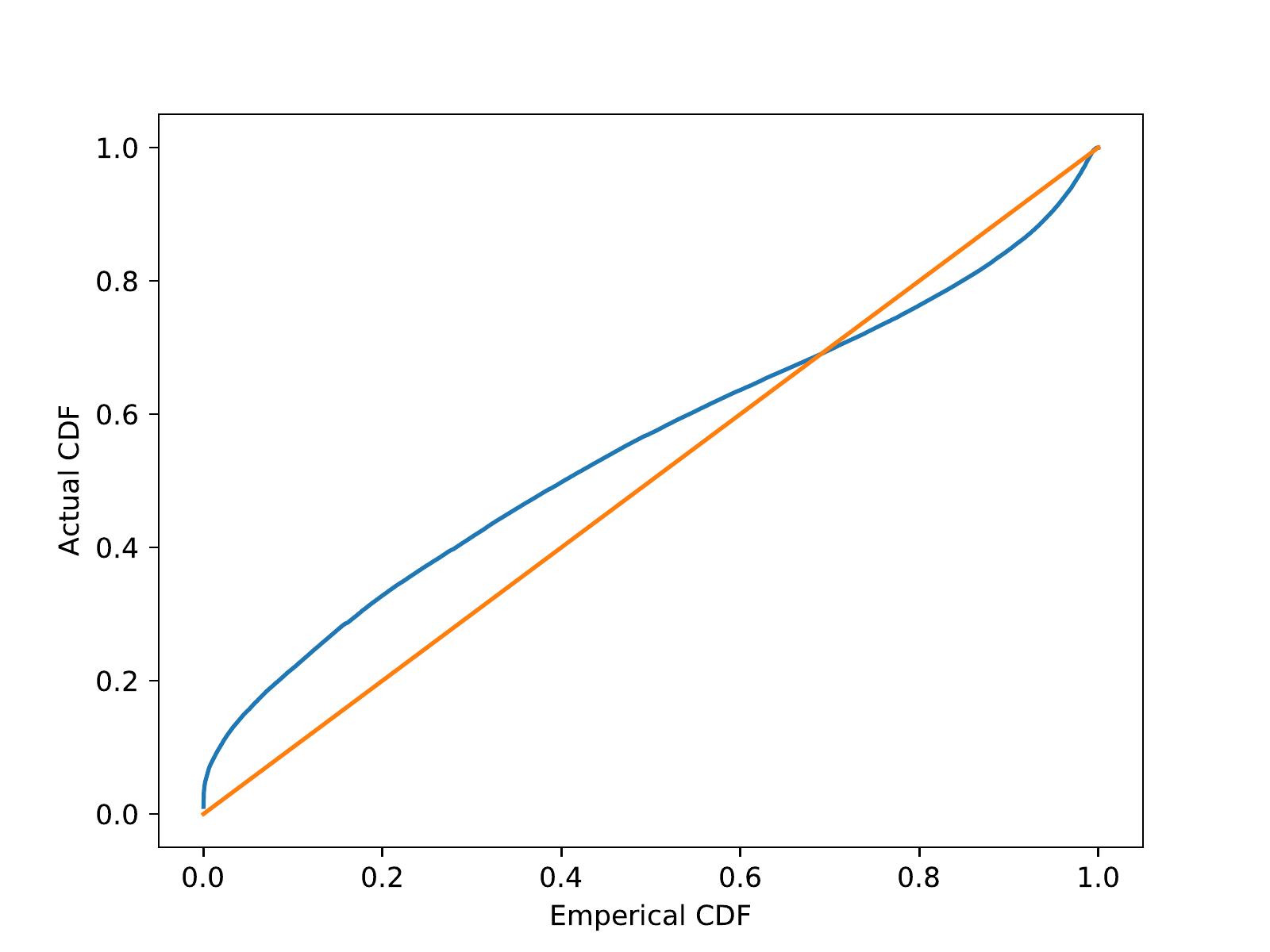}
  \caption{StarCraft II}
  \label{fig:sfig2}
\end{subfigure}
\begin{subfigure}{.25\textwidth}
  \centering
  \includegraphics[width=\linewidth, trim={0.5cm 0.1cm 1.6cm 1.4cm}, clip]{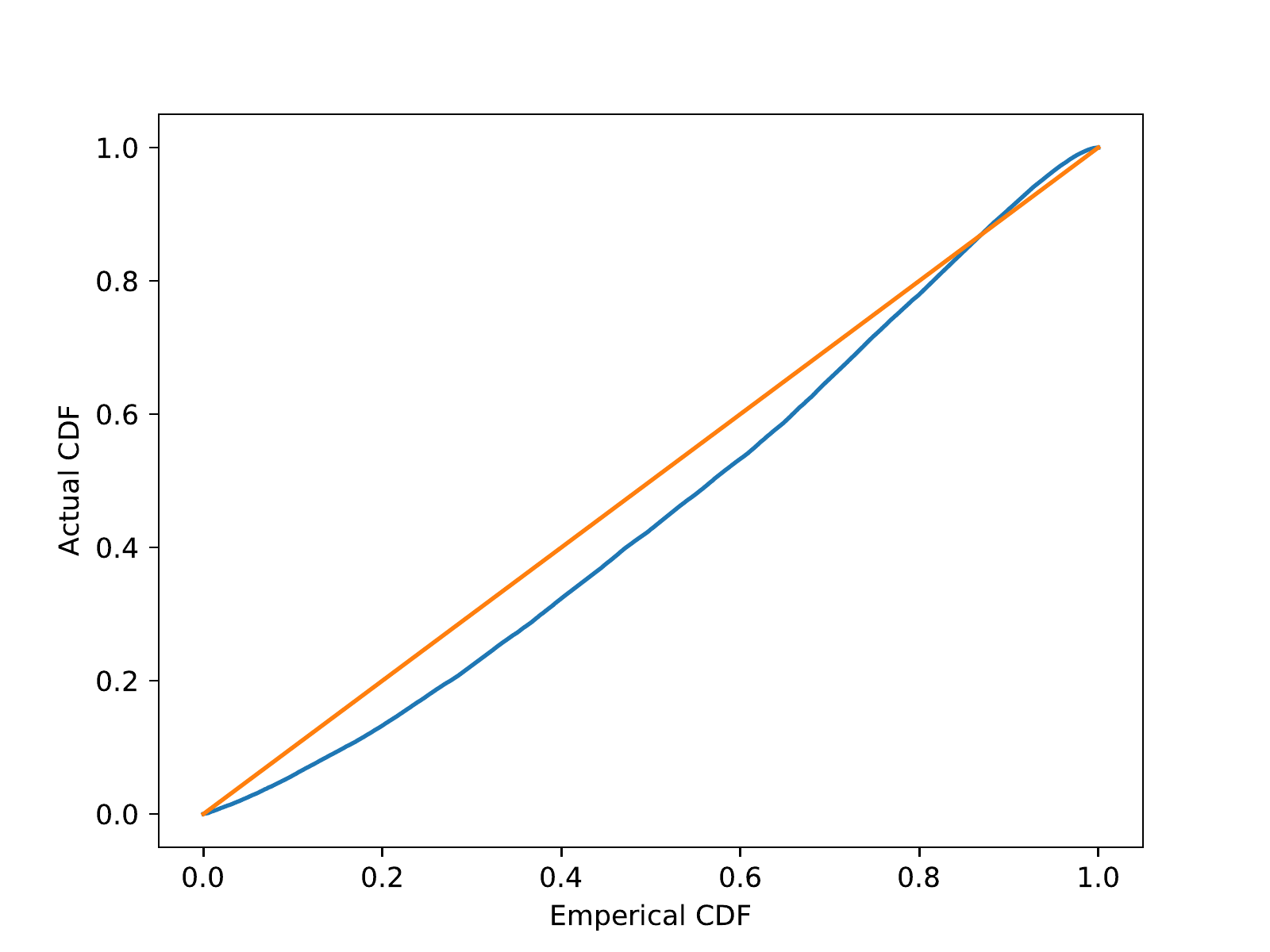}
  \caption{StackOverflow}
  \label{fig:sfig1}
\end{subfigure}%

\caption{P-P plots for various datasets for this work (above) and THP (below).}
\label{fig:PPplotsGRN}
\end{figure*}

We would like to verify that the model describes the structure present in the data accurately however it is difficult to work with non-stationary and history dependent distributions present in real world data. The Time-Rescaling Theorem \cite{Papangelou1974} states that any point process with a conditional intensity function can be mapped to a Poisson Process with unit parameter. 
Alongside the Time-Rescaling theorem, we can use the learnt conditional intensity function to obtain transformed variables $z_j$ which are independent and should be exponentially distributed with rate 1. Since the cumulative distribution function (CDF) of the target distribution is known, we can use P-P plots to measure the deviation from this ideal behaviour. P-P plots plot the empirical CDF with the actual CDF which should be a straight line equally inclined to both the axes. 
Fig. \ref{fig:PPplotsGRN} shows the P-P plots for our Recurrent Graph Network (RGN) (above) and Transformer Hawkes Process (THP) (below) for three different datasets. It can be seen that for all the three cases, the P-P plot for THP has substantially larger deviations from the expected straight line whereas Recurrent Graph Network (RGN) is remains close to the straight line. This shows that RGN better learns the structure in the data compared to state-of-the-art THP.
\subsection{Model Interpretability}
The ability to incorporate structure using Recurrent Graph Network helps in making the deep learning model more interpretable. The formulation of our model allows us to see the event-class dependency i.e how a certain class $i$ depends on a certain class $j$ whenever an event occurs allowing us to study the dynamics of inter-class influence as the event stream is observed. Moreover, incorporating multi-headed mechanism allows the model to attend to different information in different subspaces. Fig \ref{attn} shows the visualization of two attention heads in the model at three different event timestamps for the StarCraft II dataset. A lighter intensity corresponds to lower attention while a more pronounced color corresponds to higher attention score to a node. 
Other transformer based models \cite{thp, sahp} have to average out the attention scores over all events to find attention over event classes thus losing out on crucial temporal information. Although the model allows us to study inter-class dependency at event timestamps, we can easily extend it to continuous time by using spline-based interpolation \cite{spline} or other interpolation techniques.

\subsection{Ablation Studies}
We conduct ablation experiments to understand the impact of number of attention heads in the multi-headed attention mechanism of Graph Attention Network. We keep the sizes of all the parameters of the model fixed, and only vary the number of attention heads. The evaluation metrics of interest are Log-likelihood, Event type prediction accuracy and Event time prediction error. In addition, we also study the case where the GAT module is completely removed to study the improvement that inclusion of the GAT module brings over a fully LSTM based model.

The result for the experiment on the StarCraft II dataset is shown in table \ref{nheads}. As shown in figure \ref{attn}, various attention heads in the model attend to different features and table \ref{nheads} confirms our hypothesis that doing so improves the performance of the model especially on the log-likelihood metric. We also notice a stark drop in model performance when the graph attention module is removed. 

\begin{table}[b]
\addtolength{\tabcolsep}{-4.1pt}
\caption{Sensitivity to number of Attention Heads}
\label{nheads}
\centering
\begin{tabular}{c | c | c | c}
\hline
\# Attention Heads & LL & Type Accuracy &  Time Error \\
\hline
0 & 1.03 & 43.95 & 1.269\\
1 & 1.38 & 47.00 & 1.261\\
2 & 1.44 & 47.15 & 1.259\\
4 & 1.49& 47.18& 1.260\\
8 & 1.50 & 47.31 & 1.254\\
\hline
\end{tabular}
\end{table}
\vspace{-5pt}
\subsection{Reduction in Computational Complexity}
One of the major drawbacks of Transformer based approaches is the quadratic order complexity in space and time. For a sequence with length $N$, the number of activations required to store the self-attention matrix is in the order of $\mathcal{O}(N^2)$. This problem is further exacerbated when Transformer layers are stacked or multiple heads are used. Another issue that arises during implementation is due to sequence padding of the mini-batch. All the sequences in the mini-batch are padded with zeros to ensure they have the same length as the longest sequence in the mini-batch. All the sequence attention matrices scale as the square of the longest sequence length in the mini-batch $\mathcal{O}( \max(N) ^ 2)$ leading to a lot of wasted memory and redundant computation.
In contrast, the attention mechanism in RGN attends over event types rather than individual events. This makes the attention matrix scale in the order of $\mathcal{O}(|\mathcal{Y}|^2)$ instead of $\mathcal{O}(\max(N)^2)$ which leads to dramatic savings in memory consumed as usually $|\mathcal{Y}| << \max(N)$. This also makes the memory consumed by model independent of the sequence length.
Table \ref{comp} shows the number of activations in the attention mechanism (in Millions) and total number of operations (MFLOP) for both the aforementioned models. We report the total number of operations performed by both models to find the historical embedding for all events when a sequence with average length is given from all the datasets. The model hyper-parameters are described in the supplementary material. We can see that RGN has lower attention activations and computations compared to THP for all datasets except for MIMIC-II where the trend is reversed. This is due to the fact that the assumption of $|\mathcal{Y}| < N$ fails to hold for small sequences of average length $4$ and a large number of event classes $75$ present in the dataset. 

Table \ref{gpumem} shows the amount of GPU memory used by the models. We can see that for the Financial Transactions dataset, only a mini-batch of size 1 can be fit into an Nvidia RTX2080Ti GPU with 11 Gigabytes of memory. No such issues were faces for RGN which has a very small footprint due to the small number of event types - $2$ in the dataset. The trend reverses for the MIMIC-II dataset where the large number of classes cause RGN to consume more memory than THP.
\vspace{-5pt}
\begin{table}[!t]
\addtolength{\tabcolsep}{-4.1pt}
\caption{Complexity Analysis}
\label{comp}
\centering
\begin{tabular}{c  c | c c}
\hline
Dataset & Model & \makecell{$\#$ Computations \\ (MFLOP)} &  \makecell{$\#$ Activations \\ (M)} \\
\hline
\multirow{2}{*}{RT} & THP & 89.88 & 0.41  \\
                            & \B{This Work} & \underline{\B{84.10}} & \underline{\B{0.17}} \\
\hline
\multirow{2}{*}{SO} & THP & 3209.67 & 3.03  \\
                            & \B{This Work} & \underline{\B{409.23}} & \underline{\B{1.83}} \\
\hline
\multirow{2}{*}{Financial} & THP & 103294.46 & 174.58  \\
                            & \B{This Work} & \underline{\B{1343.42}} & \underline{\B{32.62}} \\
\hline
\multirow{2}{*}{MIMIC II } & THP & \underline{2.00} & \underline{0.017}  \\
                            & \B{This Work} & \B{23.21} & \B{0.081} \\
\hline
\multirow{2}{*}{SC II} & THP & 1621.84 & 1.49  \\
                            & \B{This Work} & \underline{\B{318.58}} & \underline{\B{1.17}} \\
%
%
\hline
\end{tabular}
\end{table}

\begin{table}[!t]
\caption{Peak GPU Memory Usage (GB).}
\label{gpumem}
\centering
\begin{tabular}{c  c | c c}
\hline
Dataset & Batch Size & THP &  \B{This Work} \\
\hline
Retweets & 64 & 2.08 & \underline{\B{1.03}}  \\
StackOverflow & 16 & 9.75 & \underline{\B{3.65}}  \\
MIMIC-II & 64 & \underline{1.63} & \B{5.07}  \\
Financial & 1 & 7.15 & \underline{\B{1.02}}  \\
StarCraft II & 16 & 10.29 & \underline{\B{3.63}}  \\
\hline
\end{tabular}
\end{table}

\section{Conclusion}
In this paper, we present a Recurrent Graph Network to learn the underlying Marked Temporal Point Process from event streams. Our key innovation was incorporating structural information using a dynamic attention mechanism over event types rather than all past events in the sequence. This leads to improved performance on log-likelihood and event prediction metrics. Moreover, it also leads to reduction in time and space complexity across almost all datasets. The model improves interpretability by allowing us to understand the influence of one event type over the others as more events are observed by the model.  We also present a new interesting benchmarking dataset for point process evaluation.

\bibliography{main}
\bibliographystyle{ieeetr}
\end{document}


\title{Learning Point Processes using Recurrent Graph Network: Supplementary Material}
\maketitle

\begin{theorem}[Time-Rescaling Theorem]
\label{pythagorean}
Let $0 < t_1 < t_2, ...,t_N < T$ be a realization from a point process with conditional intensity function $\lambda(t | \mathcal{H}_t)$, define
\begin{equation}
    z_j = \int_{t_{j-1}}^{t_j} \lambda(t | \mathcal{H}_t) \; dt
\end{equation}
for $j=1,...,N$ and $t_0 = 0$. Then, $z_j$ are independent exponentially distributed random variables with rate parameter 1.
\end{theorem}

\section{Appendix A}
\section{Monte Carlo Estimate of the Integral}
\begin{align}
    \hat{\Lambda}_{\MC} = \sum_{j=1}^{L_i} (t_{j} - t_{j-1} ) \left( \frac{1}{N_\tau} \sum_{k=1}^{N_\tau} \lambda(\tau_k) \right)
\end{align}

where, $\tau_k \sim \textup{U}(t_{j-1}, t_j)$ and $N_\tau$ is the number of points used for evaluation. This allows for an unbiased estimate of the integral as $\mathbb{E}(\hat{\Lambda}_{\MC}) = \Lambda$ (Proof in supplementary materials). Similarly, the gradients can be approximated using:
\begin{align}
        \nabla \hat{\Lambda}_{\MC} = \sum_{j=1}^{L_i} (t_{j} - t_{j-1}) \left( \frac{1}{N_\tau} \sum_{k=1}^{N_\tau} \nabla \lambda(\tau_k) \right)
\end{align}

\label{unbiasedproof}
\begin{align}
    \hat{\Lambda}_{\MC} & = \sum_{j=1}^{L_i} (t_{j} - t_{j-1}) \left( \frac{1}{N_\tau} \sum_{k=1}^{N_\tau} \lambda(\tau_k) \right) \\
    \mathbb{E}[\hat{\Lambda}_{\MC}] & = \sum_{j=1}^{L_i} (t_{j} - t_{j-1}) \left( \frac{1}{N_\tau} \sum_{k=1}^{N_\tau} \mathbb{E}[\lambda(\tau_k)] \right)
\end{align}
As $\tau_k \in (t_{j-1}, t_j)$ are i.i.d, they have the same expectation $\mathbb{E}[\tau_j]$.
\begin{align}
    \mathbb{E}[\hat{\Lambda}_{\MC}] & = \sum_{j=1}^{L_i} (t_{j} - t_{j-1}) \left(\mathbb{E}[\lambda(\tau_j)] \right) \label{exp}
\end{align}
The support of $\mathbb{E}[\tau_j]$ lies in $(t_{j-1}, t_j)$,
\begin{align}
    \mathbb{E}[\lambda_j] = \int_{t_{j-1}}^{t_j} \frac{1}{(t_j - t_{j-1})} \lambda(s) \: ds
\end{align}
substituting in Eq. \ref{exp},
\begin{align}
    \mathbb{E}[\hat{\Lambda}_{\MC}] & = \sum_{j=1}^{L_i} (t_{j} - t_{j-1}) \left( \int_{t_{j-1}}^{t_j} \frac{1}{(t_j - t_{j-1})} \lambda(s) \: ds \right)
\end{align}
simplifying,
\begin{align}
    \mathbb{E}[\hat{\Lambda}_{\MC}] & = \sum_{j=1}^{L_i} \int_{t_{j-1}}^{t_j} \lambda(s) \: ds \\
    \mathbb{E}[\hat{\Lambda}_{\MC}] & = \int_{0}^{T} \lambda(s) \: ds = \Lambda \;\;\;\;\;\;\;\;\; \qed
\end{align}
Thus, the Monte Carlo approximation is unbiased.

\begin{table}[!t]
\addtolength{\tabcolsep}{-4.1pt}
\caption{Dataset statistics}
\label{dataset}
\centering
\begin{tabular}{c | c c c c c}
\hline
Dataset & $|\mathcal{Y}|$ & Seq. Length &  \multicolumn{3}{c}{\# of Seq.} \\
\cline{3-6}
 &  & Mean & Train & Val & Test\\
\hline
Retweets & 3 & 109 & 20000 & 2000 & 2000 \\
StackOverflow & 22 & 72 & 4777 & 530 & 1326 \\
MIMIC-II & 75 & 4 & 527 & 58 & 65 \\
Financial & 2 & 2074 & 90 & 10 & 100 \\
StarCraft II & 16 & 78 & 6141 & 1316 & 1317 \\
\hline
\end{tabular}
\end{table}

\section{Implementation Details} \label{implementation}
Dropout \cite{dropouts} probability of $0.1$ is used to reduce overfitting. We also make use of Layer Normalization in the input-to-graph update and edge update stages to prevent explosion of gradients and activations to stabilize training. ADAM is used as the optimizer \cite{adam} and the learning rate is chosen to be $10^{-4}$.
The final complete loss function to be minimized is: 
\begin{align}
    \mathcal{L} = \mathcal{L}_\lambda + \beta_y \mathcal{L}_y + \beta_t \mathcal{L}_t
\end{align}
In the loss function $\mathcal{L}$, $\beta_y$ is set to 1 while $\beta_t$ is set to 100 to scale the L2 error.
The hyper-parameters for various datasets are shown in table \ref{hyperparam}.
The learning rate was $10^{-4}$ for all the datasets and the optimizer used was ADAM \cite{adam}. LeakyReLU parameter was set to $\alpha = 0.2$. The model was trained till $50$ epochs for all the datasets. We stack 2 layers of GAT for graph-to-graph update. Additionally, we add a residual connection that allows for skipping through the GAT Layer for better flow of gradients 


\begin{table}[!h]

\addtolength{\tabcolsep}{-4.1pt}

\caption{Hyper-parameters}
\label{hyperparam}
\centering
\begin{tabular}{c | c | c | c | c }
\hline
Dataset & $d_{in} = d_v = d_u$ & $d_e$ &\# Heads & TBPTT steps\\
\hline
Retweets & 256 & 16 & 8 & 20\\
StackOverflow & 256 & 16 & 16 & 20\\
Financial & 256 & 16 & 16 & 20\\
MIMIC II & 128 & 16 & 8 & 20\\
StarCraft II & 256 & 32 & 8 & 40\\
\hline
\end{tabular}
\end{table}

The hyper-parameters for THP were the ones prescribed by the authors and the official github implementation was used for results. For StarCraft II, the model parameters were $M = 256$, $K = V = 64$, $d_{inner} = 2048$, $\#$ Heads = $8$.

For RMTPP, NHP and THP, instead of performing $\int_{t_j}^{\infty} t \: p(t | \mathcal{H}_t) \: dt$ to find the time of the next event, we add a FC layer to the models to directly predict the time of the next event $\hat{t}$ using the latent embeddings.
We modify RMTPP to use the same $\lambda^*$ expression as ours instead of the original one proposed by the authors as it restricted the conditional intensity expression to the exponential family for a tractable likelihood computation. This also allows for a fair comparison between the models. This is not done for NHP because NHP allows for latent embeddings to be generated at any time $t$, thus we have used the one proposed by the authors.

\begin{figure*}[!t]
\centering
\includegraphics[width=0.9\textwidth, trim={0cm 0cm 0cm 0cm}, clip]{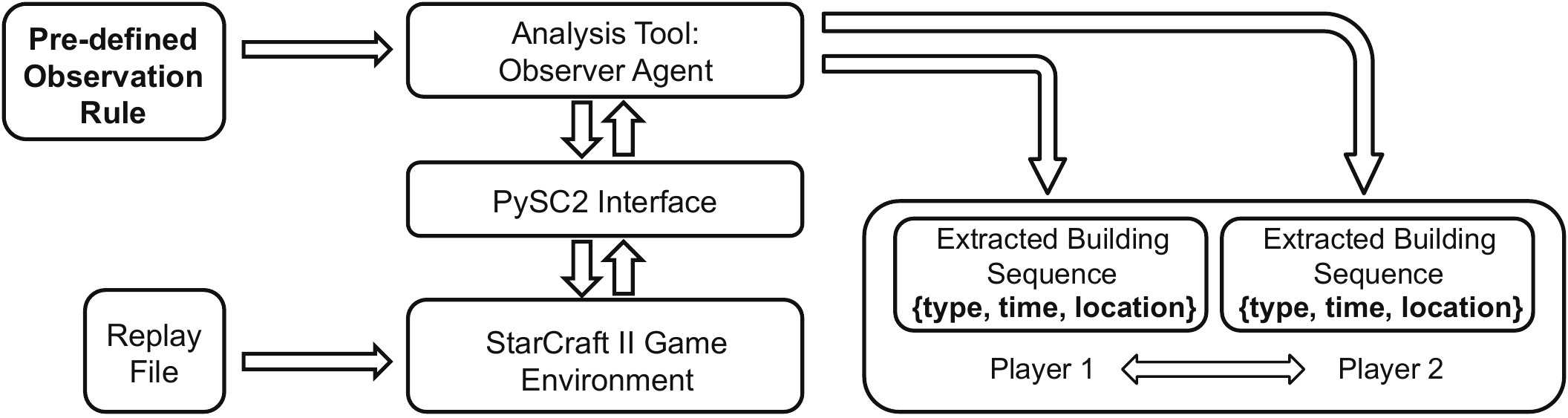}
\caption{Data Generation Tool}
\label{sc2_extract}
\end{figure*}

\begin{figure*}[!t]
\centering
\includegraphics[width=\textwidth, trim={0cm 0cm 0cm 0cm}, clip]{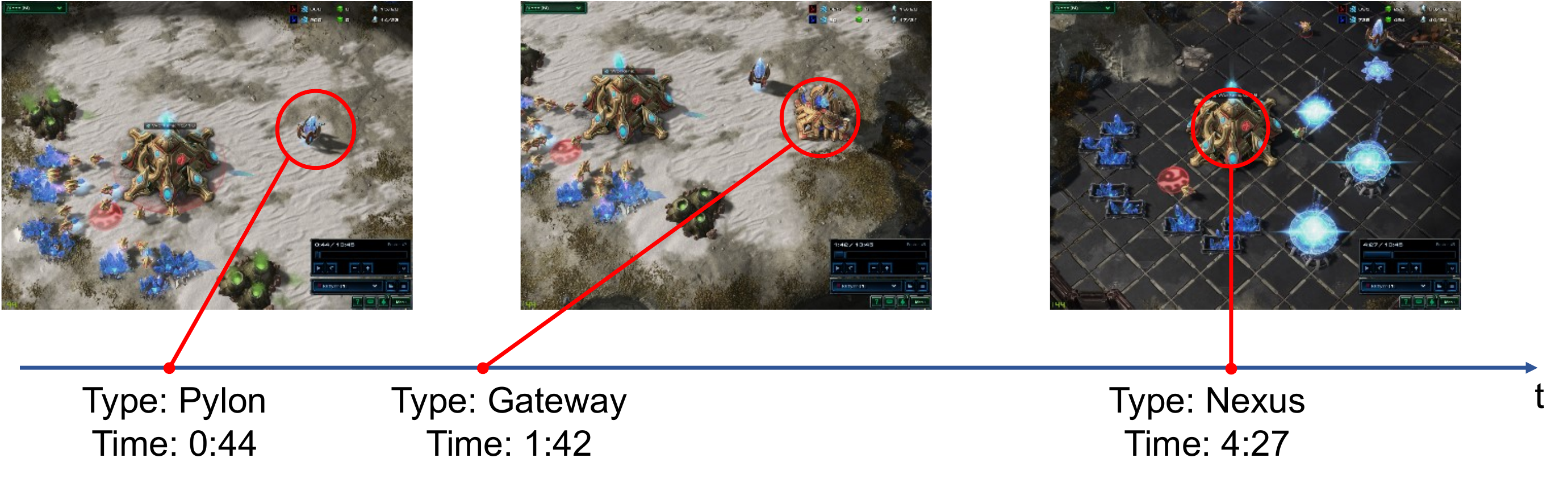}
\caption{Example Sequence}
\label{game}
\end{figure*}

\section{StarCraft II Dataset}\label{AppSC2}

StarCraft II is a real time strategy game in which players play against each with the objective of building an army and defeating the opponent. The vast event space of the game leads to significant complexity and has led to the game becoming a popular AI testbed. The large amount of available human replay files make it possible to create a suitable dataset for this work. To achieve dataset creation, we build a data generation tool based on the open source PySC2 environment and the StarCraft II machine learning API supported by the game developer. The generation tool runs the original game environment and an observer agent that gathers data from the game environment through the PySC2 interface. It performs data extraction based on specific observation rules defined by the user to generate data accordingly. As shown in Fig.~\ref{sc2_extract}, for the dataset used in this work, we define an observation rule that records certain player actions as event sequences from observing the game environment running the replay files of one versus one games. 

There are different types of buildings in StarCraft II and each serves different purpose. For example, there are buildings that produce army and buildings that develop technology. The location of buildings also has important influences. For example, players can choose to put buildings in clusters in order to more effectively protect them or they can choose to put buildings in separate locations so that the opponent has more difficulties in finding them. 
Therefore, the strategy that a player adopts can be largely reflected in the building construction information. In addition, the evolution of strategy taken by each player has a dynamic that depends on interaction with its opponent. The correlation between two players' s actions thus contains important information about their strategies.

Based on the preceding discussion, for data used in this work, we choose to extract sequences of building constructions. To achieve this, the observation rule is defined with a list of action-of-interests related to building constructions. The observer agent extracts building construction actions of both players in the game. It records three dimensions of each actions: building type, building time and building location. Each extracted data point contains two sequences: one for each player. 

A part of a game replay used in the event-to-event sequence is illustrated in Fig.~\ref{game}: the player construct a Pylon type building at the beginning, which provides resources for building more units. This is followed by a Gateway building for army production at time of 1:42. Later in the game, the player build a Nexus, which is used for resource gathering, at a remote location. This move is performed to expand the occupied territory. It is also worth noting that the shown visualization is for demonstration purpose only, and the data extraction tool as described does not rely on the fully rendered game. It extracts information from the API directly to improve data generation efficiency.

\bibliography{aaai22}
\bibliographystyle{ieeetr}